\documentclass[french]{hermes-journal}

\usepackage{todonotes}
\usepackage{lscape}
\usepackage{rotating}

\DeclareMathOperator*{\argmin}{arg\,min}
\DeclareMathOperator*{\argmax}{arg\,max}

\hsp{2014}{x}{y}

\firstpagenumber{1}

\title{Rendu basé image avec contraintes sur les gradients}

\subtitle{15 juin 2016}

\author[1]{Grégoire}{Nieto}
\author[2]{Frédéric}{Devernay}
\author[3]{James}{Crowley}

\address{Laboratoire Jean Kuntzmann, Univ. Grenoble Alpes, Inria Grenoble Rhône-Alpes, 655, avenue de l'Europe, F-38330 Montbonnot-Saint-Martin, France.}
        {gregoire.nieto@inria.fr}
\address{Laboratoire Jean Kuntzmann, Univ. Grenoble Alpes, Inria Grenoble Rhône-Alpes, 655, avenue de l'Europe, F-38330 Montbonnot-Saint-Martin, France.}
        {frederic.devernay@inria.fr}
\address{Laboratoire d’Informatique de Grenoble, Univ. Grenoble Alpes, Inria Grenoble Rhône-Alpes, 655, avenue de l'Europe, F-38330 Montbonnot-Saint-Martin, France.}
        {james.crowley@inria.fr}

\abstract{Multi-view image-based rendering consists in generating a novel view of a scene from a set of source views.
In general, this works by first doing a coarse 3D reconstruction of the scene, and then using this reconstruction to establish correspondences between source and target views, followed by blending the warped views to get the final image. Unfortunately, discontinuities in the blending weights, due to scene geometry or camera placement, result in artifacts in the target view. In this paper, we show how to avoid these artifacts by imposing additional constraints on the image gradients of the novel view. We propose a variational framework in which an energy functional is derived and optimized by iteratively solving a linear system. We demonstrate this method on several structured and unstructured multi-view datasets, and show that it numerically outperforms state-of-the-art methods, and eliminates artifacts that result from visibility discontinuities.}

\resume{Le rendu basé image consiste à générer un nouveau point de vue à partir d'un ensemble de photos d'une scène. On commence en général par effectuer une reconstruction 3D approximative de la scène, utilisée par la suite pour synthétiser l'image cherchée à partir des images sources. Malheureusement, les discontinuités dans les poids des images sources, dues à la géométrie de la scène ou au placement des caméras, causent des artefacts visuels dans la vue résultante. Dans cet article nous montrons qu'une façon d'éviter ces artefacts est d'imposer des contraintes supplémentaires sur le gradient de l'image synthétisée. Nous proposons une approche variationnelle suivant laquelle l'image cherchée est solution d'un système linéaire résolu de façon itérative. Nous testons la méthode sur plusieurs jeux de données multi-vues structurés et non-structurés, et nous montrons que non seulement elle est plus performante que les méthodes de l'état de l'art, mais elle élimine aussi les artefacts créés par les discontinuités de visibilité.}


\motscles{Rendu basé image, reconstruction 3D, imagerie computationelle.}

    {
        \bigskip\noindent\hspace*{-1cm}%
        \parbox{1cm}{\LARGE\ding{46}}\begin{minipage}{\linewidth plus 1pt minus 1pt}\rule{\linewidth}{0.5pt}\\[4pt]
    }{
        \newline\rule{\linewidth}{0.5pt}
        \end{minipage}
        \medskip
    }

\makeatletter
\DeclareRobustCommand\onedot{\futurelet\@let@token\@onedot}
\def\@onedot{\ifx\@let@token.\else.\null\fi\xspace}

\makeatother

\begin{document}

\maketitle

\newpage

\section{Extended Abstract} \label{sec:abstract}
Multi-view image-based rendering consists in generating a novel view of a scene from a set of source views. In general, this works by first doing a coarse 3D reconstruction of the scene, and then using this reconstruction to establish correspondences between source and target views. The final image can be obtained by warping and blending the input views~: this is a direct rendering method. An alternative approach is to use a variational method, in which a smoothness term is used to regularize the solution. The expression of the energy is derived from the Bayesian formulation of the posterior~: it is composed of a data term that forces the solution to fit the input images in intensity, and a smoothness term that accounts for what we know \textit{a priori} about the solution, assumed to be a relatively smooth natural image. This energy functional is derived and optimized by iteratively solving a linear system.

The recent work of \citeNP{pujades_bayesian_2014} derived new weights of contribution of each input image in the energy, that are based on the reconstructed scene geometry. Unfortunately, discontinuities in the these weights, due to imperfect geometry reconstruction or camera placement, result in artifacts in the target view. In this paper, we show how to avoid these artifacts by imposing additional constraints on the image gradients of the novel view. We propose a variational framework in which a new data term is added to the energy functional. This term forces the solution to be close to the data (the input views) in the gradient domain. It comes from the observation that a contour in the input images that is visible from the target view, should also appear in the image solution.

Our rendering method is generic, and could be applied to any camera placement. In particular we provide details on the derivation of the warps and the weights of contribution based on a set of depth maps that are generated from the input views thanks to an Multi-View Stereo algorithm. We demonstrate our variational rendering algorithm on several structured (\emph{HCI Light Field Benchmark Datasets} and \emph{Stanford Light Field Archive}) and unstructured \cite{strecha_benchmarking_2008} multi-view datasets, and show that it numerically outperforms state-of-the-art methods, and eliminates artifacts that result from visibility discontinuities. Moreover it enforces the \emph{continuity} property that was first stated by \citeNP{buehler_unstructured_2001} as a property that any ideal algorithm should satisfy. 

\newpage

\section{Introduction}
\label{sec:introduction}

Le rendu basé image consiste à générer un nouveau point de vue à partir d'un ensemble de photos d'une scène. Il s'opère généralement en deux étapes successives: une pahse de reconstruction de la géométrie de la scène et une phase de synthèse de vue. On commence par effectuer un modèle 3D approximatif de la scène, appelé \emph{proxy géométrique}. Un \emph{proxy géométrique} est obtenu par un logiciel de MVS (\emph{Multi-view Stereo}), et son format peut-être un nuage de points, un ensemble de cartes de profondeurs ou un maillage. Il permet d'estimer une fonction permettant d'aller des coordonnées en pixel dans l'image du point de vue cible aux coordonnées dans une image source, par projection du point de l'image cible sur le \emph{proxy} puis reprojection dans l'image source (figure~\ref{fig:multiview}). Cette fonction est utilisée conjointement aux images sources pour synthétiser le point de vue voulu par une méthode directe ou variationnelle. 

Le récent travail de \citeNP{pujades_bayesian_2014} propose une formulation bayésienne du problème de rendu basé image, construite sur le travail précédent de Wanner et Goldluecke~\cite{goldluecke_superresolution_2009, wanner_spatial_2012}. C'est une méthode variationnelle, c'est-à-dire une méthode qui propose d'estimer l'image cherchée par minimisation d'une fonction de coût, une énergie. Chaque pixel de la solution est le résultat de la contribution, directe ou indirecte, de plusieurs pixels des vues sources. Ces contributions sont pondérées par des termes qui apparaissent dans l'expression de l'énergie. Ils ont montré que les poids des pixels des images sources dans l'énergie pouvaient se déduire formellement des propriétés de la caméra, du contenu de l'image, et de la précision du \emph{proxy géométrique}, amenant une nouvelle formalisation des heuristiques de poids proposées initialement par \citeNP{buehler_unstructured_2001}. La plupart des << propriétés désirables qu'un algorithme idéal de rendu basé image devrait avoir >>~\cite{buehler_unstructured_2001} eurent alors une explication formelle, sauf la propriété de \emph{continuité}. En effets les contributions des vues sources varient brutalement d'un pixel de la vue cible à l'autre, soit parce que la limite du champ de vue d'une caméra est atteinte ou qu'elle est occultée (sa contribution tombe à 0), soit parce que les contributions ne sont pas lisses au sein même de la même image source, puisque l'estimation d'un \emph{proxy géométrique} est bruitée et que le calcul des poids des contributions repose la reconstruction du \emph{proxy}.

Dans cet article, nous montrons qu'une façon d'éviter ces artefacts est d'imposer des contraintes supplémentaires sur le gradient de l'image synthétisée. Ces contraintes viennent d'une simple observation: les contours d'image dans la vue cible doivent aussi être des contours dans les images sources où ces parties sont visibles. Une fonctionnelle d'énergie similaire à celle de \citeNP{pujades_bayesian_2014} est développée, composée de l'habituel terme sur les données (\emph{data term}) et un terme de régularisation (\emph{smoothness term}), mais le terme sur les données contient un terme additionnel qui prend en compte les contraintes sur les gradients. Nous montrons que tenir compte à la fois de l'intensité et du gradient dans les méthodes de rendu basé image apporte une solution élégante au renforcement de la propriété de \emph{continuité} initialement énoncée par \citeNP{buehler_unstructured_2001}.

\section{Travaux antérieurs}
\label{sec:related_work}

\subsection{Rendu basé image (IBR)} 
Les techniques de rendu basé image ont été décrites et classifiées par \citeNP{shum_image-based_2008}. La plupart des méthodes de l'état de l'art~\cite{kopf_image-based_2013,sinha_image-based_2012,lipski_correspondence_2014,lipski_virtual_2010,chaurasia_depth_2013} utilisent une reconstruction de la géométrie de la scène plus ou moins précise, appelée géométrie intermédiaire ou \emph{proxy géométrique}. \citeNP{ortiz-cayon_bayesian_2015} proposent de segmenter les images en super-pixels et de calculer la qualité de plusieurs algorithmes d'IBR afin de choisir le meilleur pour chaque super-pixel. Toutes ces techniques sont inspirées par la méthode directe de \citeNP{buehler_unstructured_2001}, qui effectuent une combinaison des $k$ plus proches vues, pondérées par les angles et les distances à la vue cible, garantissant ainsi un champ de mélange lisse. La continuité du mélange résultant dans le domaine image est certifiée par le renforcement du lissage spatial de ces poids, mais des artefacts temporels sont toujours présents si l'ensemble des caméras qui contribuent est trop éparse. \citeNP{davis_unstructured_2012} proposent une technique de rendu par subdivisions des points de vue sources pour créer un meilleur champ de mélange. Néanmoins, les poids des contributions sont encore calculés d'après des règles heuristiques et le choix des caméras pour le rendu est totalement arbitraire. 

\subsection{But d'une approche variationnelle}
Le but d'une approche variationnelle comme celle de \citeNP{wanner_spatial_2012} est d'estimer une image $u$ -- une fonction qui à tout pixel, ou point image 2D, renvoie une couleur -- à partir des données, les $k$ images sources appelées $v_k$. L'estimateur $\hat{u}$ de la solution u doit maximiser la probabilité \emph{a posteriori} d'observer l'image cherchée $u$ sachant nos données en entrée $v_k$ et la probabilité \emph{a priori} de $u$~: on l'appelle estimateur MAP (\emph{Maximum a posteriori}). Ne sachant pas calculer cette probabilité \emph{a posteriori} nous l'exprimons autrement à l'aide du théorème de Bayes en fonction de la vraisemblance et des probabilités \emph{a priori}. En prenant nos probabilités indépendantes et normalement distribuées, on arrive à la conclusion que la solution $\hat{u}$ doit minimiser la somme de deux termes $E_{color}$ et $E_{prior}$. La méthode est aussi qualifiée d'inverse, car ne sachant pas exprimer la solution en fonction des données, on va plutôt chercher à exprimer les données en fonction de la solution, supposée connue.

\subsection{Se débarrasser des heuristiques} 
Se débarrasser des heuristiques et du réglage manuel des paramètres est une idée clé de \citeNP{wanner_spatial_2012}. La contribution de chaque vue dans l'estimation de la solution est automatiquement déduite d'équations mathématiques. \citeNP{pujades_bayesian_2014} vont plus loin en intégrant l'incertitude géométrique dans le formalisme bayésien. Ils obtiennent alors de nouveaux poids qui favorisent les caméras satisfaisant à la fois la cohérence épipolaire (\emph{epipole consistency}) et la déviation angulaire minimale (\emph{minimal} \emph{angular} \emph{deviation}), deux principes établis par  \citeNP{buehler_unstructured_2001} pour décrire l'algorithme d'IBR idéal. Cependant leur méthode n'offre pas de cadre formel pour satisfaire le principe de continuité (\emph{continuity principle}), en particulier près des limites du champ de vue des caméras. Nous montrons qu'introduire un terme additionnel dans la fonctionnelle énergie, qui contraint non seulement les intensités mais aussi les gradients de la solution, apporte une solution élégante au principe de continuité.

\subsection{Rendu en haute résolution}
L'idée clé pour un rendu en haute résolution est que la qualité de l'image solution dépend souvent de la contrainte de l'espace de recherche. Par conséquent, trouver la bonne régularisation ou \emph{a priori} sur la solution est une question cruciale pour l'obtention d'images d'excellente qualité. La contribution principale de \citeNP{fitzgibbon_image-based_2003} est l'utilisation d'\emph{a priori} calculés à partir de grandes bases de données de textures afin de contraindre la solution, indépendamment des données relatives à la scène. Cette idée fut récemment étendue par \citeNP{flynn_deepstereo_2015} qui effectuent une synthèse de nouvelle vue à partir d'un réseau de neurones, entraîné par une gigantesque base d'images prises tout autour du monde. Au contraire notre méthode ne repose pas sur de forts \emph{a priori} sur la nouvelle image à synthétiser, mais tente plutôt de mieux exploiter les données procurées par les images sources pour ajouter de nouvelles contraintes sur la solution. En somme notre algorithme ne requiert pas ces énormes bases de données pour produire des images de haute qualité. Cependant les deux approches ne sont pas incompatibles, et on pourrait imaginer intégrer notre méthode à celle de \citeNP{flynn_deepstereo_2015}.

\subsection{Fusion d'images dans le domaine du gradient}
La fusion d'images dans le domaine du gradient a reçu beaucoup d'intérêt ces dernières années, en commençant par l'article phare de \citeNP{perez-poisson_2003}, pour des applications dans l'édition d'images~\cite{mccann_real-time_2008}, l'\textit{inpainting}~\cite{levin_learning_2003} ou les panoramas~\cite{agarwala_interactive_2004, zomet_seamless_2006}. Le travail le plus proche du notre en rendu basé image est probablement celui de \citeNP{kopf_image-based_2013}, qui effectue un rendu de gradient, suivi d'une intégration pour produire la couleur de l'image. Néanmoins cette méthode se limite l'interpolation de points de vue, et elle ne traite pas des configurations génériques, dans lesquelles les points de vue sources sont très différents et non structurés. Notre méthode s'attelle à ce problème en proposant un cadre plus générique pour le rendu basé image multi-vues.

\begin{figure}[fig:multiview]{La reconstruction 3D de la scène permet de mettre en correspondance la vue cible $u$ avec les vues sources $v_k$.}
\centering
\includegraphics[width=0.90\columnwidth]{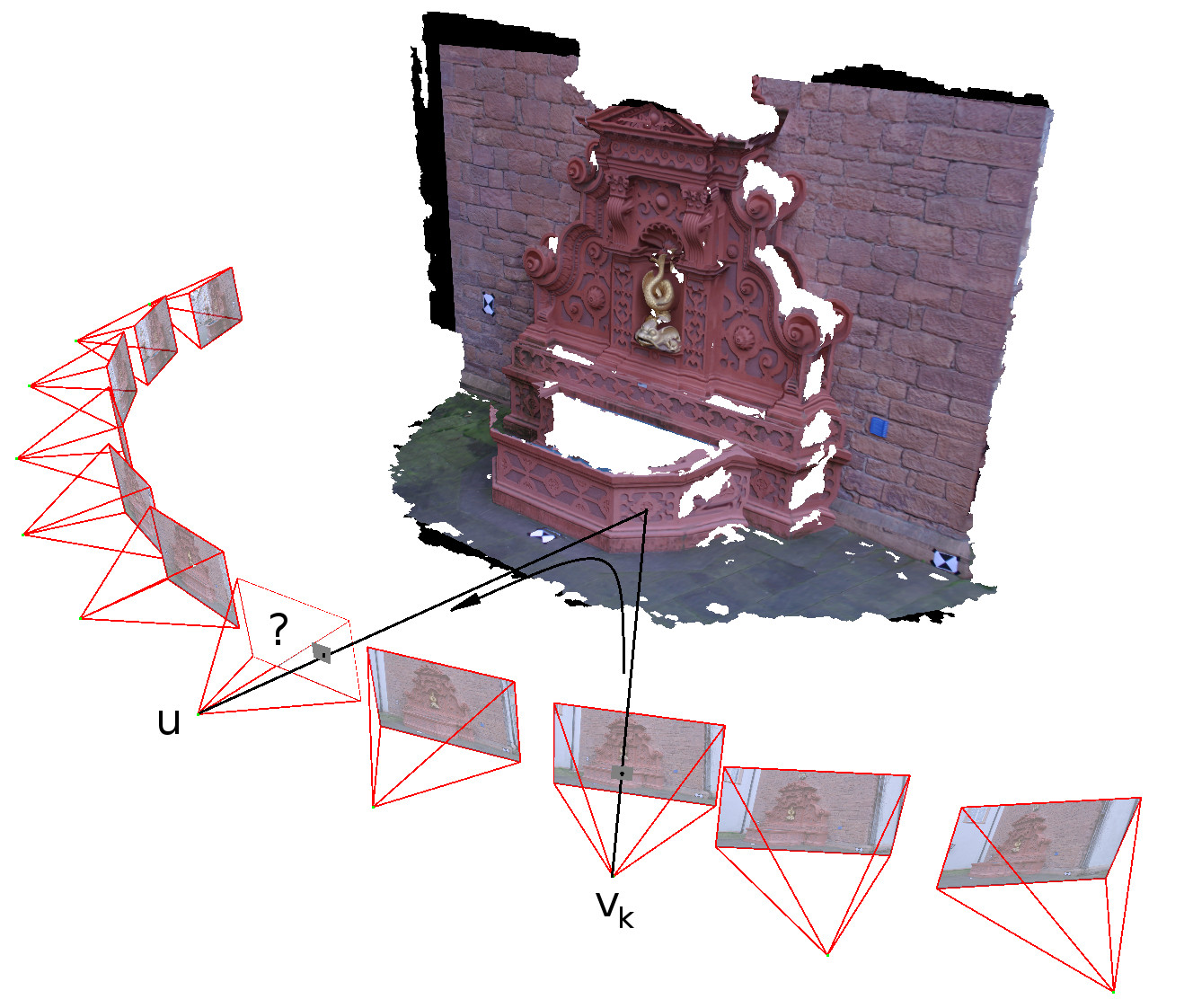}
\end{figure}

\section{Aperçu général de l'approche}
\label{sec:overview}

L'objectif de notre méthode est de synthétiser une nouvelle image optimale $u$ au point de vue cible à partir des images sources $v_k$. Les images sont des fonctions définies sur des sous-ensembles continus de $\mathbb{R}^2$ qui renvoient une intensité ou une couleur~: $u: \Gamma \subset \mathbb{R}^2 \rightarrow \mathbb{R}$ et $v_k: \Omega_k \subset \mathbb{R}^2 \rightarrow \mathbb{R}$. Dans un souci de simplicité, les valeurs des images sont supposées scalaires, mais il serait aisé de généraliser aux images couleurs $v_k: \Omega_k \rightarrow \mathbb{R}^3$. Dans toutes nos expériences nous traitons les images dans l'espace couleur RGB. La discrétisation de ces fonctions est abordée dans la section \ref{sec:dicret}. 

Notre méthode de rendu-basé image est décomposable en deux étapes indépendantes~: la reconstruction 3D (section~\ref{sec:depth}) et le rendu (section~\ref{sec:model}). 

\textbf{Reconstruction 3D}
Le but de cette étape est recaler les images sources via les \emph{warps} $\tau_k$ qui transforment tout point ${\bf x}_m = (x_m, y_m)$ de la vue source $k$ en son correspondant ${\bf x}_p = (x_p, y_p)$ dans la vue cible~:
\begin{equation}
  \begin{array}{c c c c}
    \tau_k: &\Omega_k      &\rightarrow   &\Gamma \\
            &{\bf x}_m  &\mapsto       &{\bf x}_p
  \end{array}
\end{equation}

Dans un premier temps les images sources sont utilisées afin de calibrer les caméras pour extraire leurs matrices de paramètres intrinsèques et extrinsèques. Puis un algorithme de stéréo multi-vues produit un \emph{proxy géométrique}, en l'occurrence une carte de profondeur par vue source, afin d'estimer les déformations d'images (\emph{warps}) pour les projeter sur la vue cible. Définies telles quelles, les fonctions de déformation ne sont pas bijectives~: des points image dans les vues sources n'ont pas de projeté dans la vue cible parce que la projection sort du champ de vue cible  et des points de la vue cible n'ont pas de projeté inverse dans une vue source à cause des auto-occultations. La gestion de ces deux types d'occultations se fait par l'estimation d'une carte de profondeur du point de vue ciblé, et permet de retirer les points occultés des domaines $\Omega_k$ et $\Gamma$. À l'issue de cette étape, les \emph{warps} définis sur les nouveaux domaines sont bijectifs. Enfin les paramètres des caméras sources, combinés aux différentes cartes de profondeur, nous permettent de calculer les poids des contributions de chaque vue dans l'étape de rendu.

\textbf{Rendu}
Une fois obtenus les poids de contribution et les correspondances par pixel avec la vue ciblée, on peut procéder à l'étape de rendu à proprement parler. Nous développons une formulation variationnelle du modèle de rendu basé image de l'état de l'art, que nous étendons par l'ajout de contraintes sur les gradients des images sources. 

Dans la section~\ref{sec:experiments}, nous testons la méthode sur plusieurs jeux de données multi-vues structurés et non-structurés, et nous montrons que non seulement elle est plus performante que les méthodes de l'état de l'art, mais elle élimine aussi les artefacts créés par les discontinuités de visibilité.

\section{Une formulation variationnelle du rendu basé image}
\label{sec:model}

\subsection{Modèle de formation de l'image}
\label{sec:dicret}

    \begin{figure}[fig:discretization]{L'aire de la projection du pixel est colorée en gris. À gauche~: Nous supposons que la transformée de la PSF est toujours carrée~; l'intensité résultante est la moyenne des valeurs de l'image cible pondérée par les coefficients $B_{k, m, p}$ -- l'intersection (en gris foncé) entre l'aire en gris et le pixel en haute résolution. À droite~: Modèle plus précis où nous supposons seulement que le \emph{warp} est localement linéaire et que la PSF transformée est obtenue en transformant chaque coin du pixel source.}
    \centering
    \includegraphics[width=1.0\linewidth]{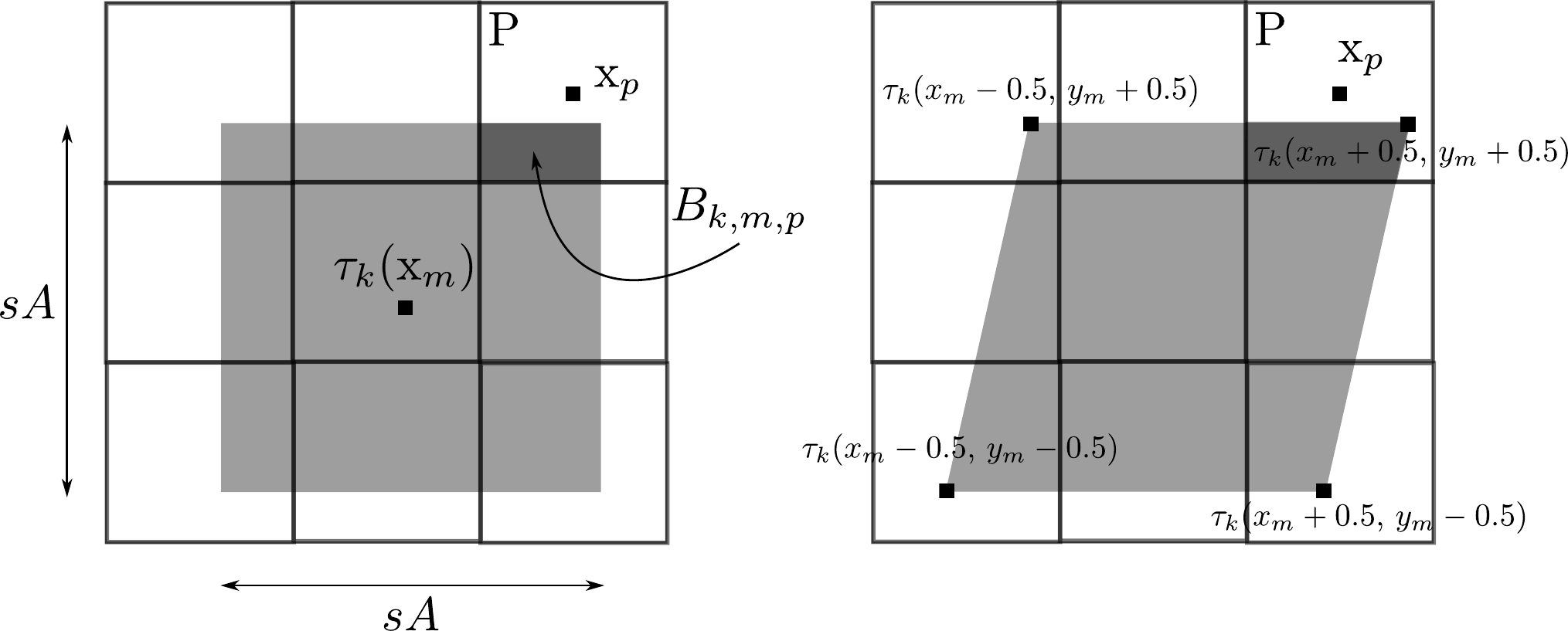}
    \end{figure}

    Comme on suppose en général dans la littérature sur la super-résolution~\cite{baker_limits_2002, hardie_joint_1997}, la valeur d'intensité $v_k({\bf x}_m)$ d'un point ${\bf x}_m$ dans l'image source $k$ peut s'écrire comme la convolution de l'image cible avec la fonction d'étalement du point ou PSF\footnote{\emph{Point Spread Function}}, notée $b$. Étant donnée une image idéale $u$ au point de vue cible, définie sur le domaine $\Gamma$, et une fonction de déformation $\tau_k$ des points de $\Omega_k$ dans $\Gamma$, si nous mettons de côté les occultations pour le moment, l'intensité de l'image observée peut s'écrire comme la relation de convolution
    \begin{equation}\label{eq:conv}
      v_k({\bf x}_m) = \int_{\Omega_k}u\circ\tau_k({\bf x})\,b({\bf x}-{\bf x}_m)\,\mathrm{d}{\bf x},
    \end{equation}
    ou plus simplement $v_k = b \ast (u \circ \tau_k)$.
    
    La PSF $b: \Omega_k \rightarrow [0, 1]$ est la densité de probabilité qui peut s'écrire $b_k: \, \Gamma \rightarrow [0, 1]$ par changement de variable ${\bf x}' = \tau_k({\bf x})$ de telle façon que
    \begin{equation}
      v_k({\bf x}_m) = \int_{\Gamma}u({\bf x}')b_k({\bf x}'-\tau_k({\bf x}_m))\,\mathrm{d}{\bf x}'.
    \end{equation}
    
    Il y a plusieurs manières de calculer la PSF transformée $b_k$, selon comment on modèle la PSF initiale $b$. L'hypothèse la plus commune est de considérer la PSF comme étant une gaussienne 2D d'espérance ${\bf x}_m$ et de covariance ${\bf \Sigma}$. Un modèle plus simple de la PSF est de se représenter un pixel carré et uniformément sensible à la lumière, la PSF étant alors une densité uniforme. En notant $A$ l'aire du pixel centré sur $(0, 0)$ dans une vue source $k$, on obtient
    \begin{equation}
      b(x, y) =  
        \begin{cases}
          \frac{1}{A^2} & \text{if} -\frac{1}{A} \leq x, y \leq \frac{1}{A} \\
          0 & \text{sinon}.
        \end{cases}
    \end{equation}
    
    Sous l'hypothèse que le \emph{warp} $\tau_k$ est localement linéaire, la PSF transformée est un parallélogramme uniformément distribué (figure~\ref{fig:discretization}). Dans ce cas, on peut faire une hypothèse encore plus forte et supposer que le \emph{warp} préserve les pixels (leur aire et leur forme), ce qui est faux en réalité mais simplifie grandement l'implémentation. Désormais nous prendrons des pixels d'aire unitaire. Puisque l'intensité est constante et égale à $u({\bf p})$ sur toute la surface du pixel ${\bf p}$ dans la vue cible, la relation de convolution (\ref{eq:conv}) ci-dessus peut être écrite comme l'ont fait \citeNP{hardie_joint_1997}~:
    \begin{equation}
      v_k({\bf x}_m) = \sum_{{\bf p} \in\, \Gamma}u({\bf p})\int_{{\bf p}}b_k({\bf x}'-\tau_k({\bf x}_m))\,\mathrm{d}{\bf x}',
    \end{equation}
    et l'intensité du pixel dans l'image source est  
    \begin{equation} \label{eq:normal}
      v_k({\bf m}) = \sum_{{\bf p} \in \Gamma}B_{k, m, p}u({\bf p}),
    \end{equation}
    où $B_{k, m, p} = \int_{{\bf p}}b_k({\bf x}'-\tau_k({\bf x}_m))\,\mathrm{d}{\bf x}'$ est l'aire de l'intersection entre la projection du pixel dans la vue cible et le pixel ${\bf p}$ de cette même vue. Si l'échantillonnage des vues de départ et d'arrivée sont les mêmes, alors les aires d'intersection sont les coefficients bilinéaires~: c'est équivalent à interpoler bilinéairement les intensités de $u$.

\subsection{Estimation du maximum \textit{a posteriori}}

Le but de l' approche variationnelle est d'estimer une image $u$ à partir des données $(v_k^*)_{k\in[1..K]}$, où $K$ est le nombre de vues sources. L'estimateur $\hat{u}$ de la solution $u$ doit maximiser la probabilité \textit{a posteriori}~; on l'appelle estimateur MAP (Maximum \textit{a posteriori})~:
	\begin{equation}
		\hat{u} = \argmax_{u} P(u | (v_k^*)_{k\in[1..K]}).
	\end{equation}
    Ne sachant pas calculer cette probabilité \textit{a posteriori} nous l'exprimons autrement à l'aide du théorème de Bayes en fonction de la vraisemblance et des probabilités \textit{a prori}. On fait l'hypothèse que les $v_k$ sont conditionnellement indépendants. Le terme $P((v_k^*)_{k\in[1..K]})$, appelé \textit{évidence}, ne dépend pas de $u$ et peut donc être retiré de l'équation~:
    \begin{equation}
		\hat{u} = \argmax_{u} \frac{P((v_k^*)_{k\in[1..K]} | u)\,P(u)}{P((v_k^*)_{k\in[1..K]})} = \argmax_{u} \prod_{k\in[1..K]} P((v_k^*) | u)\,P(u).
	\end{equation}
	
    Le terme de vraisemblance $P((v_k^*)_{k\in[1..K]} | u)$ est la probabilité d'obtenir les données (images sources) supposant connue la solution $u$. Elle s'exprime comme le produit des probabilités $P(v_k^* | u)$ dont la loi est prise normale~: $P(v_k^* | u) \propto e^{-E_{\mathrm{color}, k}(u)}$ $E_{\mathrm{color}, k(u)}$ est un terme aux moindres carrés qui représente la somme des écarts aux images sources en terme d'intensité (couleur du pixel). Il est aussi appelé terme d'attache aux données dans la littérature~:
    \begin{equation} \label{eq:e_color_k}
      E_{\mathrm{color}, k}(u) = \frac{1}{2}\int_{\Omega_k}\omega_k(u)\left(b\ast (u\circ \tau_k) - v^*_k\right)^2\,  \mathrm{d}{\bf x}.
    \end{equation}
	Le termes $\omega_k(u)$ sont les contributions par pixel de chaque image source. Ils dépendent du gradient de la solution courante $u$ et de l'incertitude sur la géométrie reconstruite. Une formule explicite de ces contributions (\ref{eq:geometrie}) est donnée dans la section~\ref{sec:depth}. En supposant que le bruit du capteur est gaussien et identique pour toutes les images, nous notons  $\sigma_{s,k}^2 = \lambda$ sa variance, une constante strictement positive. La vraisemblance étant le produit des $P(v_k^* | u)$, elle s'exprime
	\begin{equation} \label{eq:e_color}
		P((v_k^*)_{k\in[1..K]} | u) \propto e^{-\frac{1}{\lambda}E_\mathrm{color}(u)}, \text{ avec } E_{color}(u) = \sum_{k=1}^KE_{\mathrm{color}, k}(u).
    \end{equation}
    
    La probabilité \textit{a priori} $P(u)$ représente notre connaissance \emph{a priori} sur l'image à synthétiser. Nous savons que cette dernière est naturelle, donc comporte peu de variations de gradient~: le signal est \emph{régulier}. Dans notre travail, nous utilisons un \textit{a priori} de variation totale \cite{goldluecke_approach_2010}, norme $L^1$. Une preuve de convergence est fournie par \citeNP{chambolle_algorithm_2004}. 
    \begin{equation} \label{eq:tv_prob}
      P(u) \propto e^{-E_{\mathrm{prior}}(u)}, \text{ avec } E_{prior}(u) = \int_{\Gamma} |\nabla u|.
    \end{equation}
    
    On rappelle que l'estimateur MAP $\hat{u}$ doit minimiser la probabilité \emph{a posteriori}~: 
	\begin{eqnarray}
		\hat{u} & = & \argmax_{u} P(u | (v_k^*)_{k\in[1..K]}) \\
				& = & \argmin_{u} -\ln P(u | (v_k^*)_{k\in[1..K]}) \\
				& = & \argmin_{u} -\ln P((v_k^*)_{k\in[1..K]} | u) -\ln P(u) \\
				& = & \argmin_{u} \frac{1}{\lambda} E_{color}(u) + E_{prior}(u).
	\end{eqnarray}
	Le paramètre $\lambda$ strictement positif permet de contrôler la prépondérance du terme de régularisation dans l'énergie. Notre solution $\hat{u}$ doit minimiser
    \begin{equation}\label{eg:energy}
      E(u) = E_{color}(u) + \lambda E_{prior}(u).
    \end{equation}

\subsection{Ajout du terme portant sur le gradient de l'image}

    \begin{figure}[fig:warpdisc]{Les discontinuités des \emph{warps} $\tau_k$ et des poids $\omega_k$ provoquent des artefacts. À gauche: la vue globale d'une scène de Strecha estimée avec l'énergie donnée par l'équation \ref{eg:energy}. En haut à droite: un zoom révèle des artefacts haute-fréquence. En bas à droite~: une carte de profondeur présentant des discontinuités dues à la visibilité à cet endroit de l'image.}
    \centering
    \includegraphics[width=0.98\columnwidth]{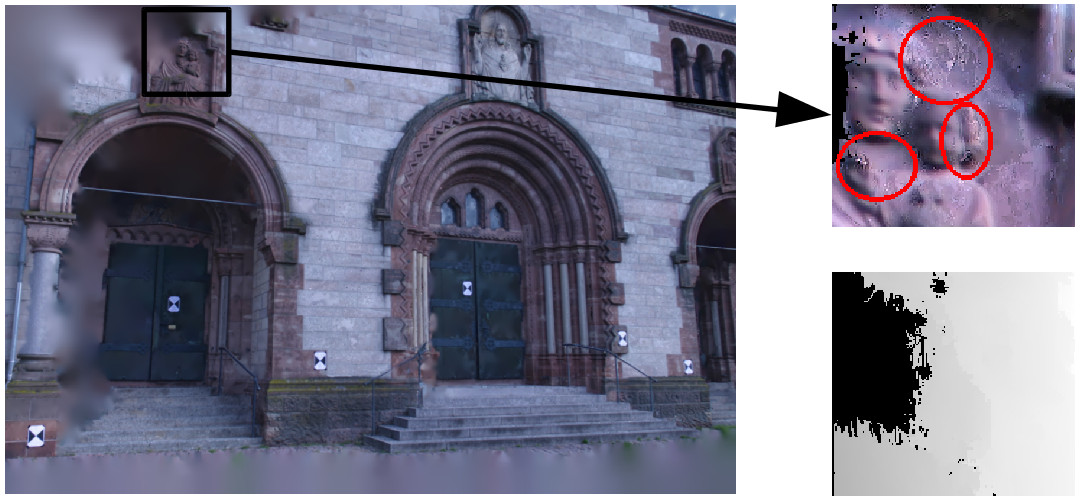} 
    \\
    \end{figure}
    
    Les cartes de profondeur estimées sont bruitées et incomplètes. Le bruit fait référence à la variance élevée et aux nombreux \emph{outliers} des carte de profondeur, dû à l'incertitude de localisation. Il peut être corrigé par un lissage ou un seuillage, au pris d'une perte de précision et d'une plus grande sparsité du modèle. Les discontinuités sont dues aux limites des champs de vue des caméras et aux auto-occultations. Par conséquent les $\omega_k(u)$ dans (\ref{eq:e_color_k}) et les $\tau_k$ peuvent être aussi bruités et discontinus, ce qui résulte en des artefacts dans la solution finale qui apparaissent comme de faux bords ou textures (figure~\ref{fig:warpdisc}). La méthode de rendu basé image devrait empêcher ces contours d'apparaître~: en fait, un contour synthétisé dans l'image solution devrait également être présent dans les images sources, là où ces parties de la scène sont visibles.
    
    Pour renforcer cette propriété, nous ajoutons un terme supplémentaire $E_{grad}(u)$, qui force la solution courante à se rapprocher des données dans le domaine du gradient. Ce terme permet en outre d'ajouter de nouvelles contraintes au système qui est alors mieux conditionné. Nous l'obtenons à partir de $P((\nabla v_k^*)_{k\in[1..K]} | \nabla u)$, la probabilité d'obtenir les gradients des images sources, sachant le gradient de l'image cible. Nous supposons que les variables aléatoires $\nabla v_k$ sont indépendantes, identiquement distribuées et obéissent à une loi normale, d'où
    \begin{eqnarray} \label{eq:energy_grad}
      E_{grad}(u) & \propto & -\ln\,P((\nabla v_k^*)_{k\in[1..K]} | \nabla u) \\
                  & \propto & -\sum_{k=1}^{K} \ln\,P(\nabla v_k^* | \nabla u) \\
                  & = & \sum_{k=1}^{K}\frac{1}{2}\int_{\Omega_k}(\nabla v_k - \nabla v^*_k)^2 \,\mathrm{d}{\bf x} \\
                  & = & \sum_{k=1}^{K}\frac{1}{2}\int_{\Omega_k}(\nabla (b\ast (u\circ \tau_k)) - \nabla v^*_k)^2\,\mathrm{d}{\bf x}.
    \end{eqnarray}
    
    Trouver $u$ qui minimise ce terme d'énergie particulier est alors équivalent à résoudre un système de $K$ équations de Laplace~:
    \begin{equation}
      \Delta (b\ast (u\circ \tau_k)-v^*_k) = 0,
    \end{equation}
    où $\Delta = \nabla.\nabla$ représente le laplacien de l'image. On en déduit alors la différentielle de la fonctionnelle~:
    \begin{equation}
      \mathrm{d}E_{grad}(u) = (|\frac{\partial \tau_k}{\partial z}|^{-1}\,\bar{b}\ast(\Delta (b\ast (u\circ \tau_k)) - \Delta v^*_k))\circ \beta_k.
    \end{equation}
    
    Les $\beta_k$ sont les déformations inverses qui apparaissent à cause du changement de variable dans l'intégrale. $\bar{b}$ est l'adjoint de la PSF $b$. Les déformations $\tau_k$ sont celles qui ont été estimées auparavant, et manquent donc de précision. Cette incertitude a un effet néfaste sur le calcul de $\Delta (b\ast (u\circ \tau_k))$. Par conséquent, nous choisissons de calculer le laplacien de $u$ d'abord, puis de le transformer dans le domaine $\Omega_k$. Sous l'hypothèse que les déformations $\tau_k$ sont localement linéaires, on peut négliger leurs dérivées au deuxième ordre et obtenir
    \begin{equation}
      \Delta (b\ast (u\circ \tau_k)) = b\ast \left(\frac{\partial \tau_k}{\partial x}^\intercal {\bf H}u \frac{\partial \tau_k}{\partial x} + 
      \frac{\partial \tau_k}{\partial y}^\intercal {\bf H}u \frac{\partial \tau_k}{\partial y}\right),
    \end{equation}
    où ${\bf H}u = \frac{\partial \nabla u}{\partial {\bf x}}$ est la hessienne de $u$.
    
    Les cartes de profondeur mal estimées causant de fortes discontinuités dans les correspondances entre les vues, la hessienne peut être très instable. Pour l'implémentation et dans ce cas seulement, nous supposons que $\tau_k({\bf x}) \approx {\bf x} + d$, de telle façon que 
    \begin{equation}\label{poor_assumption}
      \Delta (b\ast (u\circ \tau_k)) = b\ast (\mathrm{trace}({\bf H}u)\circ \tau_k) = b\ast (\Delta u \circ \tau_k).
    \end{equation}
    La forme finale de l'énergie à minimiser est donc 
    \begin{equation}\label{eg:energy_final}
      E(u) = \alpha E_{color}(u) + \gamma E_{grad}(u) + \lambda E_{prior}(u).
    \end{equation}

\subsection{Discrétisation}

    A ce stade il est légitime de se demander comment passer d'une formulation continue du problème à une solution numérique. Pour chaque pixel ${\bf m}$ de chaque vue source $k$ nous obtenons une équation similaire à (\ref{eq:normal}). Soit $\mathbf{V}^*$ le vecteur de tous les pixels de toutes les vues sources mis dans une seule grande colonne $(v_0(0), v_0(1), ..., v_{K-1}(M-1))$, $\mathbf{U}$ le vecteur colonne contenant la solution courante $(u(0), ..., u(N-1))$, et $\mathbf{B}$ la matrice $KM \times N$ qui contient les coefficients $B_{k, m, p}$. On peut donc écrire naturellement $\mathbf{V} = \mathbf{B}\mathbf{U}$. Par conséquent on exprime l'énergie (\ref{eq:e_color}) comme un système linéaire:
    \begin{equation} 
      E_{color}(\mathbf{U}) = (\mathbf{B}\mathbf{U}-\mathbf{V}^*)^\intercal \mathbf{W}(\mathbf{B}\mathbf{U}-\mathbf{V}^*),
    \end{equation}
    où $\mathbf{W}$ est une matrice $KM\times KM$ diagonale qui contient les poids $|J_{{\bf x}'}(\beta_k)|\omega_k$. Pour minimiser cette énergie nous dérivons le système linéaire, et obtenons les équations normales qui nous apportent un estimateur de la solution $\mathbf{\hat{U}}$~:
    \begin{equation} 
      \mathbf{B}^\intercal\mathbf{W}\mathbf{B}\mathbf{\hat{U}} = \mathbf{B}^\intercal\mathbf{W}\mathbf{V}^*.
    \end{equation}
    La matrice $\mathbf{B}^\intercal\mathbf{W}\mathbf{B}$ n'est en général pas inversible. Le système linaire peut être résolu par n'importe quelle méthode des moindres carrés linéaires.
    
    De même que le terme sur les couleurs de l'image, le terme portant sur les gradients est 
    \begin{equation} 
      E_{grad}(\mathbf{U}) = (\mathbf{B}\nabla\mathbf{U}-\nabla\mathbf{V}^*)^\intercal(\mathbf{B}\nabla\mathbf{U}-\nabla\mathbf{V}^*)
    \end{equation}
    et se dérive identiquement. Nous minimisons la fonctionnelle (\ref{eg:energy_final}) ainsi discrétisée via FISTA (\emph{Fast Iterative Shrinkage Thresholding Algorithm})~\cite{beck2009}.

\section{Reconstruction 3D}
\label{sec:depth}

La plupart des méthodes de rendu basé image utilisent une reconstruction 3D approximative de la scène appelée \emph{proxy géométrique}. Nous avons opté pour une représentation en cartes de profondeur car elles sont un bon compromis entre précision et exhaustivité de reconstruction. En effet, la reconstruction d'un nuage de points à l'aide d'un algorithme de l'état de l'art~\cite{furukawa_accurate_2010} est économe en données et très précise mais les données sont éparses. D'autre part, si une reconstruction de surface~\cite{kazhdan_poisson_2006,fuhrmann2014floating} est faite à partir du nuage de points dans le but de densifier les correspondances entre les vues, la précision de la géométrie diminue. Les cartes de profondeur offrent en outre l'avantage d'établir immédiatement les correspondances $\tau_k$ entre tout point ${\bf x}_m$ d'une vue source $v_k$ son projeté ${\bf x}_p$ sur la vue cible $u$.

\subsection{Coordonnées homogènes}

On notera ${\bf \bar{x}} = (x, y, 1)$ le point image ${\bf x} = (x, y)$ auquel on a rajouté une troisième coordonnée, en unités pixel. De la même façon on note ${\bf \tilde{x}} = z.{\bf \bar{x}} = z.(x, y, 1)$ le point homogène associé. Le passage des coordonnées homogènes en coordonnées euclidiennes se fait par le biais de la fonction de normalisation $N_e$ telle que $N_e({\bf \tilde{x}}) = {\bf x}$. Elle divise un point en coordonnées homogènes par sa dernière composante, ici la profondeur orthogonale $z$ du point 3D correspondant. Le passage de coordonnées euclidiennes en coordonnées étendues 3D se fait via la fonction $N_h$ telle que $N_h({\bf x}) = {\bf \bar{x}}$ par ajout d'une troisième composante. La jacobienne de $N_h$ est constante, en revanche celle de $N_e$ dépend des coordonnées du point homogène. Ces jacobiennes de normalisation en coordonnées homogènes sont données par \citeNP{heuel_uncertain_2004} à la page 110. Nous invitons le lecteur à consulter cet ouvrage pour de plus amples informations sur la géométrie projective dans le contexte de la propagation d'incertitude. 

\subsection{Calibration des caméras}

Nous utilisons une partie du logiciel de reconstruction 3D multi-vues  MVE\\~\cite{fuhrmann2014mve}. Dans un premier temps, nous corrigeons la distorsion radiale des caméras et les calibrons à l'aide d'openMVG~\cite{moulon2013bibliotheque}. Le modèle sténopé est alors choisi pour représenter les caméras: une matrice $3 \times 3$ de paramètres intrinsèques ${\bf K}_k$, ainsi que la matrice $3 \times 3$ de rotation ${\bf R}_k$ et le vecteur de translation 3D ${\bf t}_k$ permettant le changement en coordonnées monde/caméra. Le centre optique de chaque caméra peut être calculé à partir des paramètres extrinsèques: ${\bf C}_k = -{\bf R}_k^\intercal {\bf t}_k$. S'il l'on appelle ${\bf X}_m$ le point 3D associé au point homogène ${\bf \tilde{x}}_m \in \Omega_k$ situé à une distance orthogonale $z_m$ de la caméra $k$. On a alors ${\bf \tilde{x}_m} = {\bf K}_k({\bf R}_k{\bf X}_m + {\bf t}_k)$ ou encore en coordonnées monde ${\bf X}_m = {\bf R}_k^\intercal {\bf K}_k^{-1} {\bf \tilde{x}_m} + {\bf C}_k$.

Nous supposons connus les paramètres de la vue à synthétiser ${\bf K}_u$, ${\bf R}_u$, ${\bf t}_u$ et ${\bf C}_u$. 

\subsection{Calcul des cartes de profondeur}

Pour chaque vue $k$, une carte de profondeur est estimée en utilisant l'algorithme de stéréo multi-vues~\cite{goesele2007multi}. Les profondeurs obtenues $h$ sont radiales -- distances euclidiennes entre un point 3D de la scène et le centre de la caméra. Nous les convertissons en profondeurs orthogonales: 
\begin{equation}\label{eq:orth}
    z_m = \frac{h_m}{\|{\bf K}_k^{-1}{\bf \bar{x}}_m\|}.
\end{equation}

En effet en coordonnées locales à la caméra $k$, le point 3D s'écrit ${\bf X}_m = {\bf K}_k^{-1}{\bf \tilde{x}}_m = z_m.{\bf K}_k^{-1}{\bf \bar{x}}_m$. Ainsi en supposant que les distances sont en valeur absolue, on obtient $h_m = \|{\bf X}_m\| = z_m.\|{\bf K}_k^{-1}{\bf \bar{x}}_m\|$ d'où l'égalité (\ref{eq:orth}).
    
De la même façon, la dérivée spatiale $h_{{\bf x}} = \frac{\partial h}{\partial {\bf x}}$ donnant l'orientation de la surface peut être convertie en 
\begin{equation}
    z_{{\bf x},m} = \frac{\partial z_m}{\partial {\bf x}} = \frac{1}{\|{\bf K}_k^{-1}{\bf \bar{x}}_m\|}(h_{{\bf x}, m}-h_m.\frac{({\bf K}_k^{-1}{\bf \bar{x}}_m)^\intercal ({\bf K}_k^{-1}[0]\,{\bf K}_k^{-1}[1])}{\|{\bf K}_k^{-1}{\bf \bar{x}}_m\|^2})
\end{equation}

où ${\bf K}_k^{-1}[0]$ et ${\bf K}_k^{-1}[1]$ représentent respectivement la première et la deuxième colonne de ${\bf K}_k^{-1}$.

Chaque carte de profondeur est filtrée par un filtre bilatéral~\cite{kopf_joint_2007} dans le but de combler les trous.

\subsection{Correspondances par pixel}

Les déformations d'images $\tau_k$ sont calculées en projetant sur la vue cible le point 3D estimé par la carte de profondeur de la vue source $k$. Le schéma suivant permet de se rendre compte des différentes opérations qui opèrent lors de la transformation d'un point image d'une caméra à l'autre:
\begin{equation}
  \begin{array}{c c c c c c c c c c c c}
    \tau_k: & {\bf x_m} & \underset{(a)}{\longrightarrow} & {\bf \bar{x}}_m & \underset{(b)}{\longrightarrow} & {\bf \tilde{x}_m} & \underset{(c)}{\longrightarrow} & {\bf X_m} & \underset{(d)}{\longrightarrow} & {\bf \tilde{x}_p} & \underset{(e)}{\longrightarrow} & {\bf x_p}
  \end{array}
\end{equation}

(a) Le point image 2D ${\bf x_m}$ (en unités pixel) est étendu en ${\bf \bar{x}}_m$ par ajout d'une troisième coordonnée valant 1 via la fonction $N_h$. (b) La multiplication par la distance orthogonale $z_m$ vue de la caméra $k$ nous donne le point homogène ${\bf \tilde{x}_m}$. (c) Un changement de repère par le biais des matrices de la caméra $k$ permet d'obtenir les coordonnées monde du point 3D ${\bf X_m = {\bf R}_k^\intercal {\bf K}_k^{-1} {\bf \tilde{x}_m} + {\bf C}_k}$. (d) Puis le point homogène ${\bf \tilde{x}_p}$ est obtenu par changement inverse dans le repère de la caméra ciblée ${\bf \tilde{x}_p} = {\bf K}_u({\bf R}_u{\bf X}_m + {\bf t}_u)$. (e) Enfin le point image ${\bf x_p}$ résultant (en unités pixels) est calculé par normalisation euclidienne $N_e({\bf \tilde{x}_p})$. Nous avons donc
\begin{equation} \label{eq:projection}
  \tau_k({\bf x}_m) = N_e({\bf K}_u({\bf R}_u(z_m.{\bf R}_k^\intercal {\bf K}_k^{-1}N_h({\bf x}_m)+{\bf C}_k)+{\bf t}_u)).
\end{equation}

\subsection{Gestion des occultations}

\begin{figure}[fig:udepth]{La carte de profondeur de la vue cible, obtenue par projection de \textit{quads} depuis les vues sources. Images de la base \emph{fountain}.}
\centering
\includegraphics[width=0.90\columnwidth]{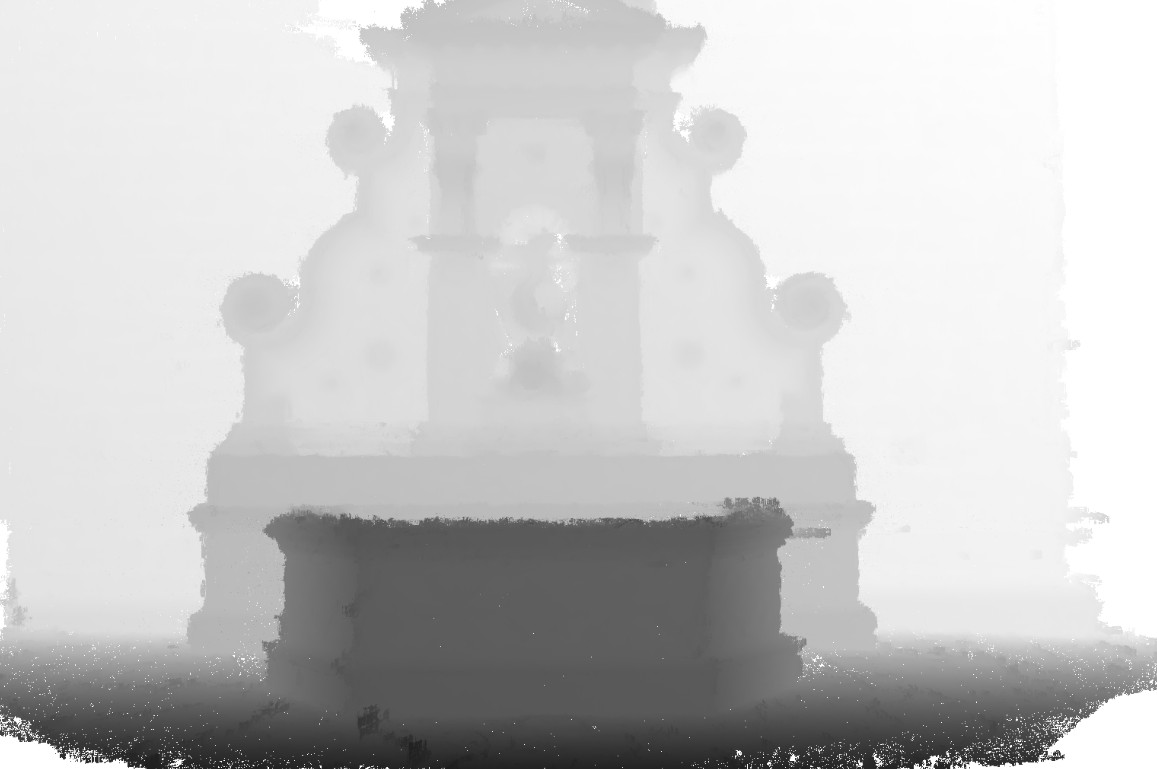}
\end{figure}

Le traitement des cartes de profondeur permet d'estimer les fonctions de déformation $\tau_k$. Telles quelles elles ne sont pas bijectives~: un point dans une vue source $k$, projeté dans la vue cible par $\tau_k$ n'est peut-être pas visible dans cette vue (auto-occultation). Il nous faut donc réduire le domaine de départ ($\Omega_k$) aux points qui sont visibles dans la vue cible~; c'est le rôle de l'étape de gestion des occultations. À l'issue de cette dernière, on a retiré des domaines de départ les pixels où les \emph{warps} ne sont pas définis à cause des occultations, les $\tau_k$ sont bijectifs sur les domaines résultants.

La gestion de la visibilité se fait en deux temps. D'abord, les pixels des vues sources sont marqués invalides si leur projeté par $\tau_k$ se situe hors des bords de l'image de la vue cible. Ensuite, la carte de profondeur $z_u$ de la vue cible est estimée pour traiter les occultations inverses (figure~\ref{fig:udepth}). À partir de chaque pixel ${\bf x}_m$ de chaque vue source $v_k$ un \textit{quad} (quadrilatère 3D) est créé à la distance estimée $z_m$ et orienté par $z_{{\bf x}}({\bf x}_m)$. Il est ensuite projeté sur la vue cible $u$ en accumulant un \textit{z-buffer} pour ne retenir que les profondeurs les plus proches. Le test de visibilité s'effectue en comparant la distance du point 3D reconstruit à la vue cible avec la profondeur estimée précédemment: si la différence se situe au-delà d'un certain seuil -- fixé arbitrairement -- alors le pixel ${\bf x}_m$ est marqué comme non visible depuis $u$. 

\subsection{Contributions des vues sources}

Les poids des contributions de chaque vue source dans le rendu sont de deux sortes~\cite{pujades_bayesian_2014}: les poids de résolution et de déformation. 

\textbf{Les poids de déformation}, donnés par le jacobien $|\frac{\partial\tau_k}{\partial {\bf x}_m}|$ de la déformation $\tau_k$, pénalisent les vues qui observent la surface de biais ou qui se situent loin de la vue cible. La jacobienne se calcule à partir des matrices des caméras, des profondeurs et des normales estimées, en dérivant en chaîne la fonction composée (\ref{eq:projection}) par rapport au point image 2D ${\bf x}_m$. Dérivons alors chacune des déformations intermédiaires énoncées dans la chaîne (\ref{eq:projection}). (e) $\frac{\partial {\bf x}_p}{\partial {\bf x}_m} = J_e({\bf \tilde{x}}_p) \frac{\partial {\bf \tilde{x}}_p}{\partial {\bf x}_m}$. (d) $\frac{\partial {\bf \tilde{x}}_p}{\partial {\bf x}_m} = {\bf K}_u{\bf R}_u \frac{\partial {\bf X}_m}{\partial {\bf x}_m}$. (c) $\frac{\partial {\bf X}_m}{\partial {\bf x}_m} = {\bf R}_k^\intercal {\bf K}_k^{-1} \frac{\partial {\bf \tilde{x}}_m}{\partial {\bf x}_m}$. (b) $\frac{\partial {\bf \tilde{x}}_m}{\partial {\bf x}_m} = z_m.\frac{\partial {\bf \bar{x}}_m}{\partial {\bf x}_m} + {\bf \bar{x}}_m z_{{\bf x},m}$. (a) $\frac{\partial {\bf \bar{x}}_m}{\partial {\bf x}_m} = J_h$. Ces équations misent bout-à-bout, on obtient
\begin{equation} \label{eq:deformation}
  \frac{\partial\tau_k}{\partial {\bf x}_m} = J_e({\bf \tilde{x}}_p){\bf K}_u{\bf R}_u{\bf R}_k^\intercal {\bf K}_k^{-1}
  \begin{pmatrix}
    z_{x,m}x_m+z_m & z_{y,m}x_m\\
    z_{x,m}y_m & z_{y,m}y_m+z_m\\
    z_{x,m} & z_{y,m}\\
  \end{pmatrix}.
\end{equation}

\textbf{Les poids de géométrie}, garantissant la \emph{déviation angulaire minimale}, c'est-à-dire pénalisant les caméras formant un angle trop grand avec la vue cible, découlent de l'incertitude de la géométrie estimée $\sigma_{g,k}^2$ et de la variance du bruit du capteur $\sigma_{s,k}^2$. Chaque vue source $k$ est ainsi pondérée par $\omega_k(u) = (\sigma_{s,k}^2 + \sigma_{g,k}^2(u))^{-1}$ avec $\sigma_{g,k}^2(u) = \sigma_{z,k}^2 (b \star (\frac{\partial \tau_k}{\partial z}^\intercal \nabla u \circ \tau_k))^2$, et $\nabla u$ le gradient de la solution courante $u$. De la même façon que pour les poids de résolution, la dérivés $\frac{\partial \tau_k}{\partial z}$ s'obtient par composition des dérivées de la chaîne (\ref{eq:projection}). (e) $\frac{\partial {\bf x}_p}{\partial z_m} = J_e({\bf \tilde{x}}_p) \frac{\partial {\bf \tilde{x}}_p}{\partial z_m}$. (d) $\frac{\partial {\bf \tilde{x}}_p}{\partial z_m} = {\bf K}_u{\bf R}_u \frac{\partial {\bf X}_m}{\partial {\bf x}_m}$. (c) $\frac{\partial {\bf X}_m}{\partial {\bf x}_m} = {\bf R}_k^\intercal {\bf K}_k^{-1} \frac{\partial {\bf \tilde{x}}_m}{\partial z_m}$. (b) $\frac{\partial {\bf \tilde{x}}_m}{\partial z_m} = z_m.\frac{\partial {\bf \bar{x}}_m}{\partial z_m} + {\bf \bar{x}}_m$. (a) $\frac{\partial {\bf \bar{x}}_m}{\partial z_m} = {\bf 0}_2$. Enfin mis bout-à-bout
\begin{equation} \label{eq:geometrie}
  \frac{\partial \tau_k}{\partial z_m} = J_e({\bf \tilde{x}}_p){\bf K}_u{\bf R}_u{\bf R}_k^\intercal {\bf K}_k^{-1}{\bf \bar{x}}_m.
\end{equation}

\section{Expériences}
\label{sec:experiments}

Afin de mettre en évidence l'influence du terme sur les gradients des images sur la qualité de la vue synthétisée, plusieurs expériences ont été conduites sur des scènes synthétiques et réelles, pour des placements de caméras structurés ou non. 

Il est important de préciser que les parties de la vue cible qui ne sont visibles par aucune des vues sources sont remplies par un algorithme de va-et-vient (\emph{push/pull} inpainting de \citeNP{gortler_lumigraph_1996}). L'effet provoqué est celui d'un remplissage par diffusion de ces zones (figures ~\ref{fig:fountain_herzjesu} et ~\ref{fig:charce_lion}).

\subsection{Base de données structurées}

Les premières expériences ont été réalisées à partir d'une base de données d'images \emph{light-field}~\cite{wanner_datasets_2013}, prises du \emph{HCI Light Field Benchmark Datasets} et de la \emph{Stanford Light Field Archive}. Pour chaque base d'images nous utilisons une matrice de vues adjacentes ($3\times 3$ ou $1\times 9$). Nous appliquons la méthode de \citeNP{wanner_globally_2012} pour estimer les disparités entre les vues et l'incertitude de la géométrie à partir des huit vues voisines. La vue centrale est rendue par chaque algorithme testé puis comparée avec l'image originale qui sert de référence pour évaluer les performances de l'algorithme. Toutes les images sources sont prises en compte dans la synthèse de point de vue.

On rappelle que $\alpha$, $\beta$ et $\gamma$ sont des réels positifs qui pondèrent chaque terme de l'énergie (17), respectivement le terme d'attache aux données sur les couleurs,  le terme d'attache aux données sur les gradients et le terme de régularisation. $\alpha$ et $\lambda$ sont fixés à leur valeur d'origine dans les expériences précédentes~\cite{wanner_spatial_2012,pujades_bayesian_2014}, respectivement 1.0 et 0.1. Nous faisons varier $\gamma$ de 0 à 3 pour observer l'influence du terme sur les gradients. Un $\gamma$ nul est bien entendu équivalent à minimiser la même fonctionnelle que celle de \citeNP{pujades_bayesian_2014}, mais notre implémentation diffère légèrement de la leur, ce qui explique pourquoi nous avons représenté les deux dans le tableau \ref{tab:results}. Nous comparons également notre méthode à celle de \citeNP{wanner_spatial_2012}. Toutes les expériences sont réalisées sur carte graphique (nVidia GTX Titan). La résolution du système prend entre 2 et 3 secondes pour des images de résolution $768\times 768$.

Le PSNR (\textit{Peak Signal to Noise Ratio}, plus il est haut mieux c'est) et le $\mathrm{DSSIM} = 10^4(1 - \mathrm{SSIM})$~\cite{wang_bovik_ssim_2004} (\textit{Structural dissimilarity}, plus il est faible mieux c'est) sont calculés par rapport à la vue de référence pour évaluer nos résultats. Le deuxième jeu d'expériences a été réalisé avec les mêmes images, mais avec une géométrie plane -- la disparité estimée est constante, ce qui correspond à un plan, le \emph{proxy géométrique} le plus grossier. L'incertitude de la géométrie est augmentée.

Notre terme sur les gradients améliore systématiquement les résultats avec une disparité estimée, et très souvent pour une disparité plane. La qualité des images générées est une fonction croissante de $\gamma$. Nous interprétons ceci par le fait que le terme sur les gradients ajoute de nouvelles contraintes au système, permettant à l'algorithme d'optimisation une meilleure convergence vers le minimum global de l'énergie.

\begin{sidewaystable}{Résultats numériques sur les bases d'images synthétiques et réelles~\cite{wanner_datasets_2013}. Notre méthode est comparée à celle de \citeNP{wanner_spatial_2012} et de \citeNP{pujades_bayesian_2014}. Le \emph{proxy géométrique} est soit estimé par \citeNP{wanner_spatial_2012} soit mis à profondeur constante avec une grande incertitude. Pour chaque \emph{light-field}, la première valeur est le PSNR (plus il est aussi mieux c'est), la seconde est $10^{-4}$ fois le DSSIM. $\mathrm{DSSIM} = 10^4(1 - \mathrm{SSIM})$~\cite{wang_bovik_ssim_2004} (plus il est faible mieux c'est). La meilleure performance est en gras. Voir la section 6.2 pour de plus amples détails sur l'expérience.}
\label{tab:results}
  \centering\footnotesize
  \begin{tabular}{ | c || c | c || c | c || c | c | c | }
    \hline
    & \multicolumn{2}{ c || }{HCI light fields, raytraced} &
                                                             \multicolumn{2}{ c || }{HCI light fields, gantry} & 
                                                                                                                 \multicolumn{3}{ c | }{Standford light fields, gantry}\\
    \hline
    & buddha & stillLife & maria & couple & truck & gum nuts & tarot \\
    \hline \hline
    Disparité estimée & & & & & & & \\
    SAVSR\cite{wanner_spatial_2012} & 42.84 \quad 17 & 30.13 \quad 58 & 40.06 \quad 53 & 26.55 \quad 226 & 33.75 \quad 408 & 31.82 \quad 1439 & 28.71 \quad 60 \\
    BVS\cite{pujades_bayesian_2014} & 42.37 \quad 18 & 30.45 \quad 55 & 40.10 \quad 53 & 28.50 \quad 178 & 33.78 \quad \textbf{407} & 31.93 \quad 1437 & 28.88 \quad 58 \\
    $\gamma = 0.0$ & 43.07 \quad 15 & 30.75 \quad 50 & 39.91 \quad 53 & 32.93 \quad 92 & 33.73 \quad 434 & 31.98 \quad 1430 & 25.37 \quad 51 \\
    $\gamma = 1.0$ & 43.28 \quad 14 & 30.83 \quad 49 & 40.14 \quad 51 & 33.05 \quad 90 & 33.82 \quad 430 & 32.08 \quad 1428 & 25.58 \quad 48 \\
    $\gamma = 2.0$ & 43.43 \quad 13 & 30.86 \quad 49 & 40.36 \quad 48 & 33.15 \quad 88 & 33.91 \quad 427 & 32.17 \quad 1426 & 25.74 \quad 46 \\
    $\gamma = 3.0$ & \textbf{43.59} \quad \textbf{12} & \textbf{30.90} \quad \textbf{48} & \textbf{40.55} \quad \textbf{46} & \textbf{33.22} \quad \textbf{87} & \textbf{33.98} \quad 423 & \textbf{32.25} \quad \textbf{1424} & \textbf{25.89} \quad \textbf{44} \\
    \hline
    Disparité plane & & & & & & & \\
    SAVSR\cite{wanner_spatial_2012} & 34.28 \quad 74 & 21.28 \quad 430 & 31.65 \quad 144 & 20.07 \quad 725 & 32.48 \quad 419 & 30.55 \quad 1403 & 22.64 \quad 278 \\
    BVS\cite{pujades_bayesian_2014} & 37.51 \quad 44 & 22.24 \quad 380 & 34.38 \quad 99 & \textbf{22.88} \quad \textbf{457} & 33.79 \quad 386 & 31.30 \quad 1378 & 23.78 \quad 218 \\
    $\gamma = 0.0$ & 37.69 \quad 42 & \textbf{22.27} \quad \textbf{377} & 34.28 \quad 100 & 22.74 \quad 468 & 34.50 \quad 367 & 31.38 \quad 1359 & 24.47 \quad 189 \\
    $\gamma = 1.0$ & 37.74 \quad 41 & \textbf{22.27} \quad 378 & 34.34 \quad 97 & 22.74 \quad 468 & 34.57 \quad 365 & 31.39 \quad 1359 & 24.51 \quad 187 \\
    $\gamma = 2.0$ & 37.80 \quad \textbf{40} & 22.26 \quad 378 & 34.40 \quad 95 & 22.74 \quad 468 & 34.66 \quad 362 & \textbf{31.43} \quad 1358 & 24.50 \quad 187 \\
    $\gamma = 3.0$ & \textbf{37.84} \quad \textbf{40} & 22.25 \quad 378 & \textbf{34.45} \quad \textbf{93} & 22.74 \quad 468 & \textbf{34.68} \quad \textbf{360} & 31.42 \quad \textbf{1357} & \textbf{24.58} \quad \textbf{185} \\
    \hline
  \end{tabular}
\end{sidewaystable}

\subsection{Base de données non structurées}

\begin{figure}[fig:fountain_herzjesu]{Résultats sur les jeux de données \emph{fountain} et \emph{herzjesu}. La colonne de droite montre un échantillon des images sources.}
\centering
\begin{tabular}{c}
\includegraphics[width=0.95\columnwidth]{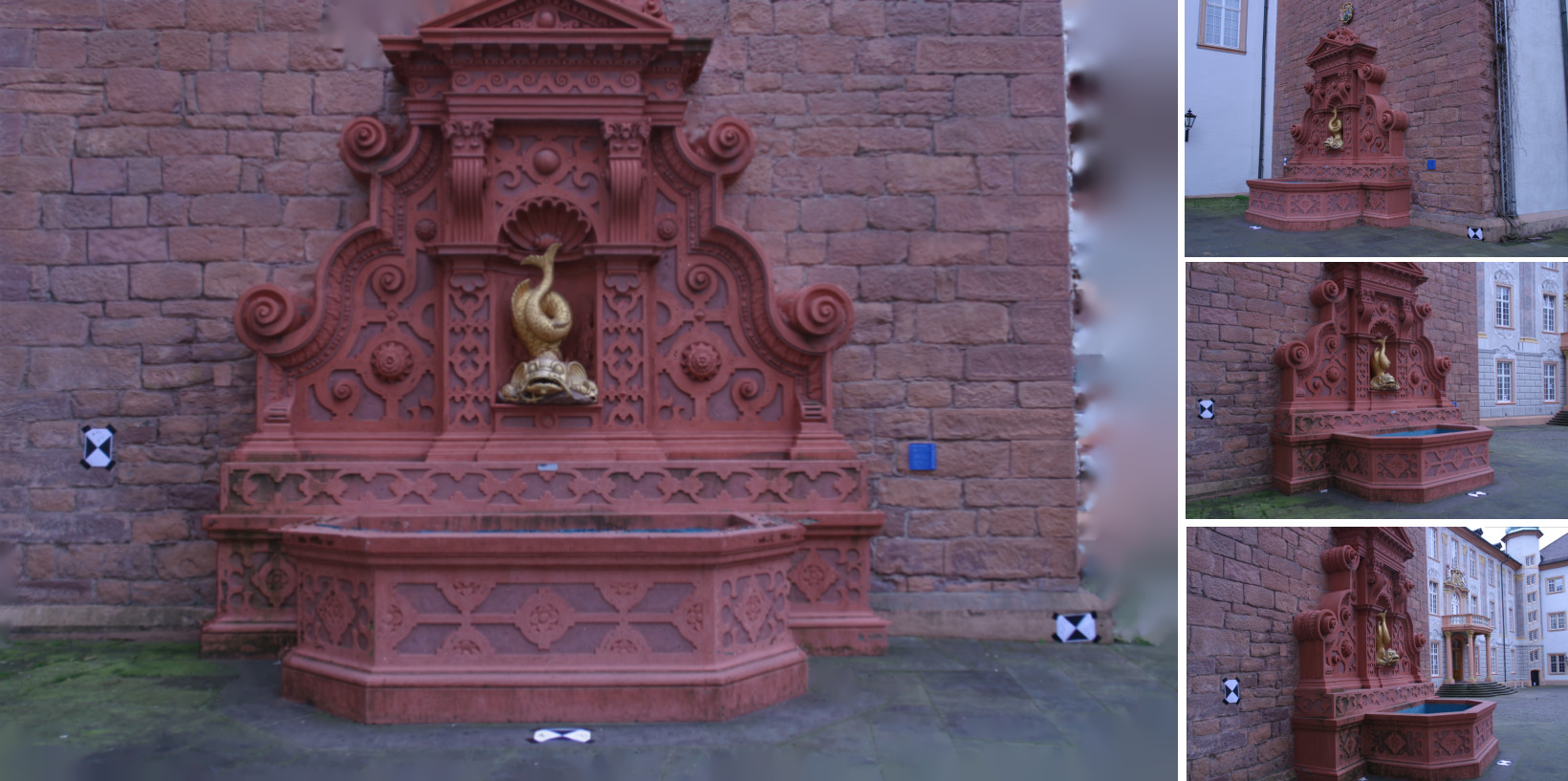} \\
\includegraphics[width=0.95\columnwidth]{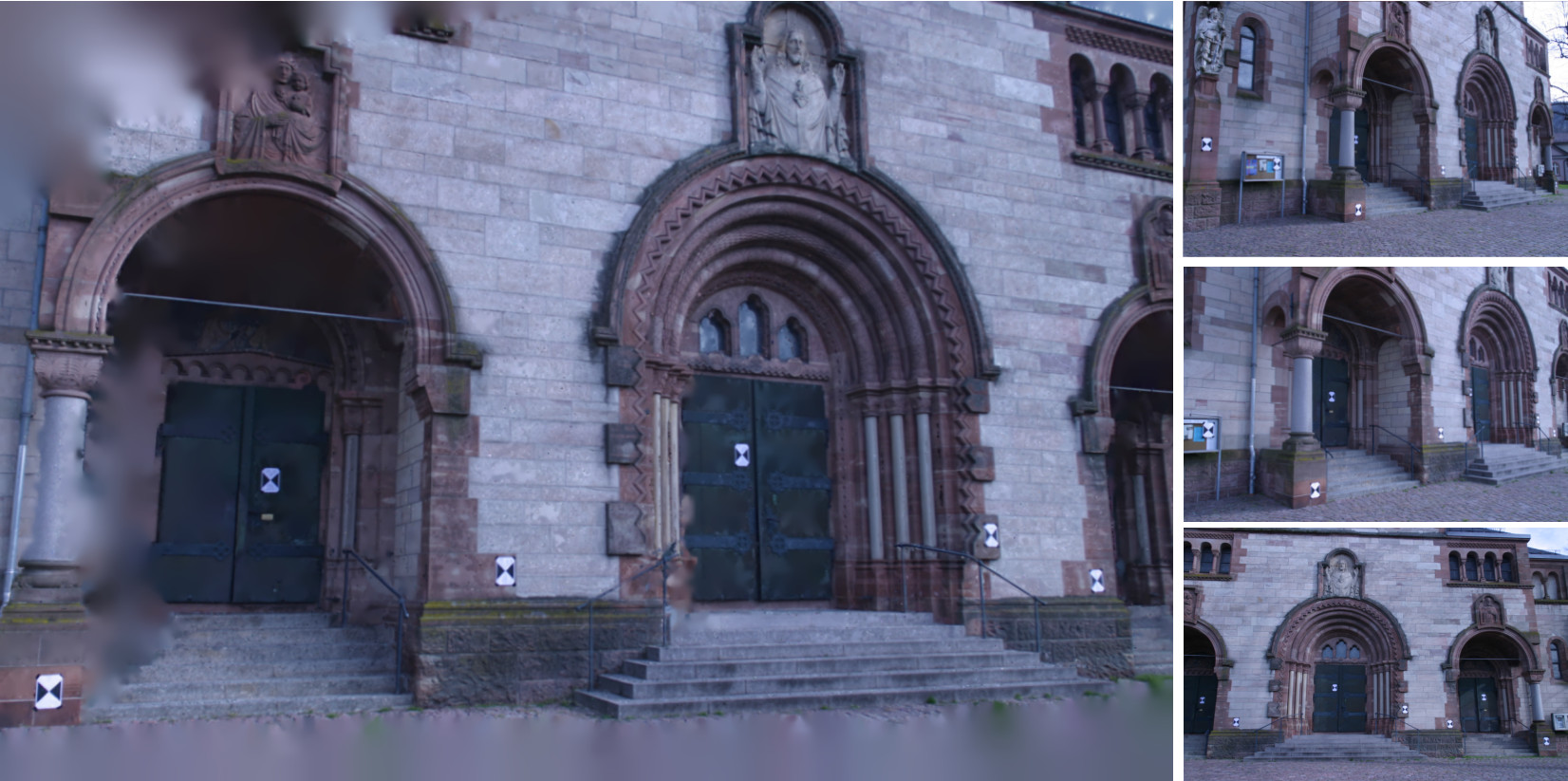} \\
\end{tabular}
\end{figure}

\begin{figure}[fig:charce_lion]{Résultats sur les jeux de données \emph{charce} et \emph{lion}. La colonne de droite montre un échantillon des images sources.}
\centering
\begin{tabular}{c}
\includegraphics[width=0.95\columnwidth]{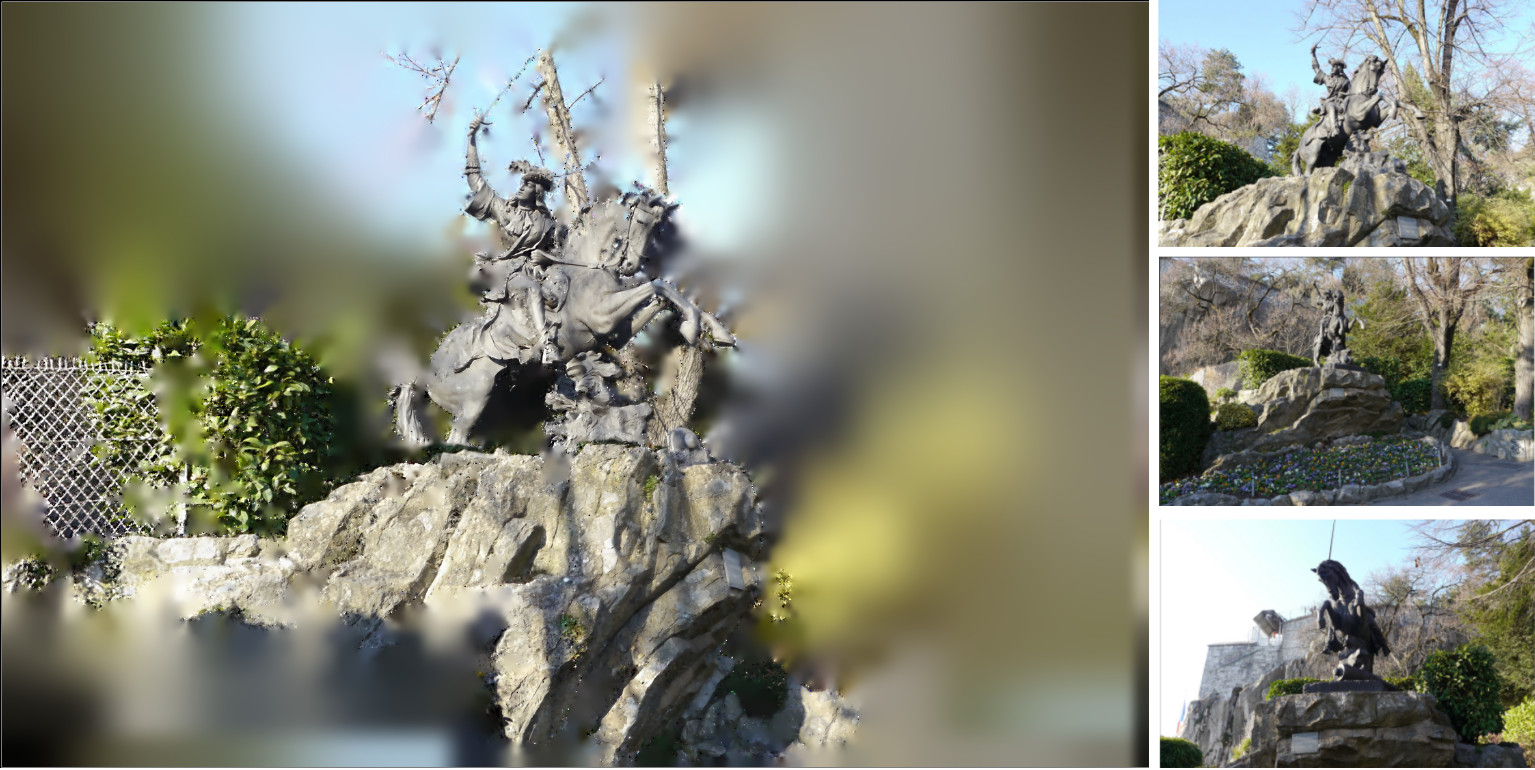} \\
\includegraphics[width=0.95\columnwidth]{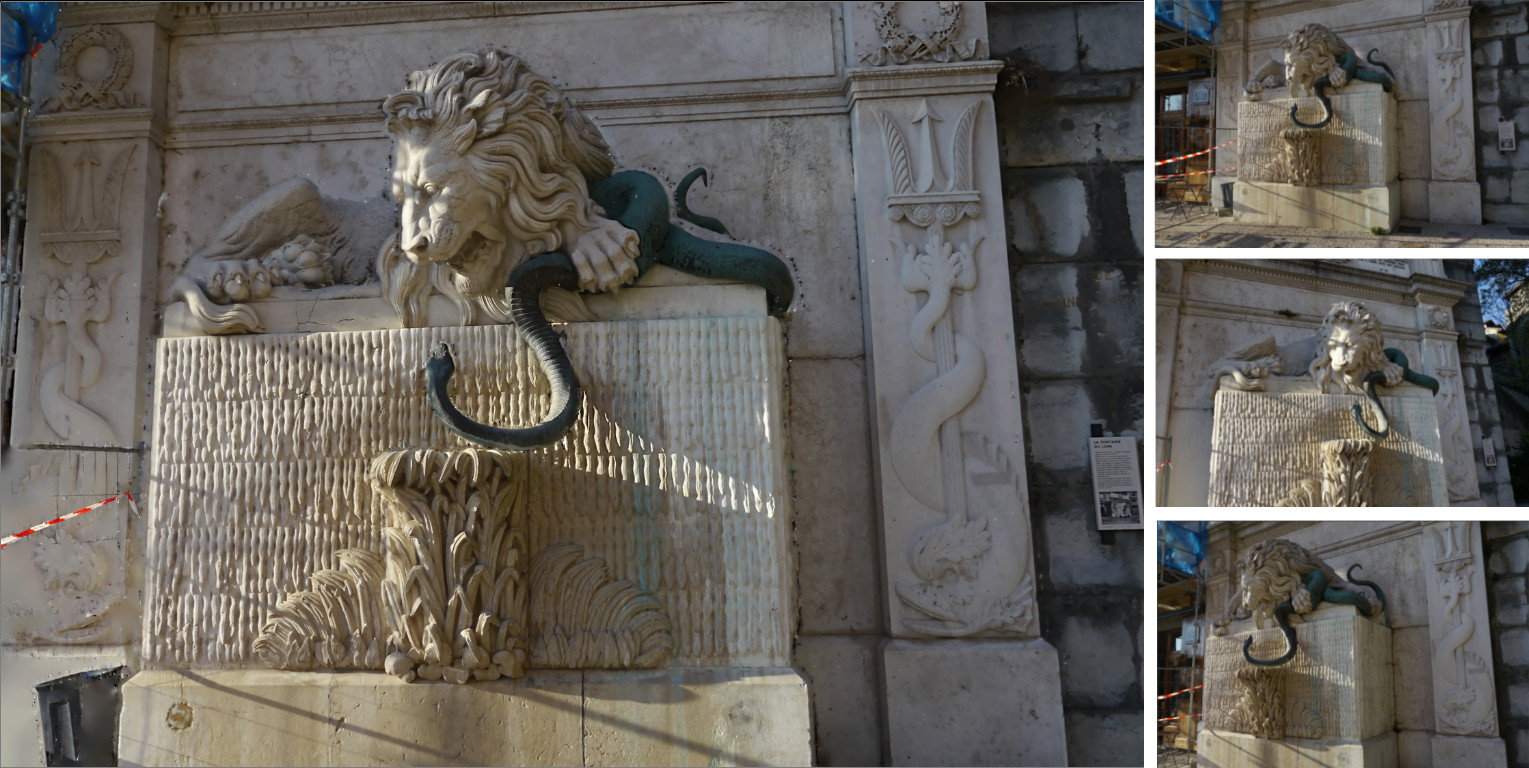} \\
\end{tabular}
\end{figure}

\begin{figure}[fig:exper]{Rendu avec différents jeux de paramètres ($\alpha, \gamma, \lambda$) qui contrôlent la proportion des différents termes dans la formule de l'énergie (\ref{eg:energy_final}). Les deux premières lignes montrent des résultats sur la collection d'images \emph{fountain}, et les deux dernières sur la collection \emph{herzjesu}. L'approche proposée est $\gamma \neq 0$.}
\centering
\begin{tabular}{cccc}
\includegraphics[width=0.25\columnwidth]{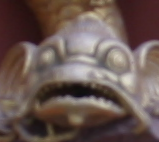} &
\includegraphics[width=0.25\columnwidth]{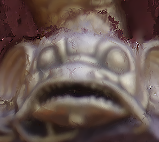} &
\includegraphics[width=0.25\columnwidth]{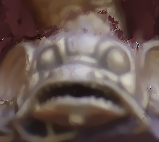} &
\includegraphics[width=0.25\columnwidth]{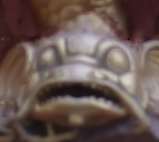} \\
&&&\\
\includegraphics[width=0.25\columnwidth]{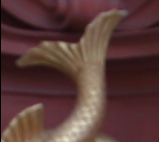} &
\includegraphics[width=0.25\columnwidth]{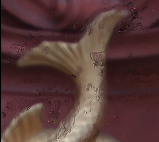} &
\includegraphics[width=0.25\columnwidth]{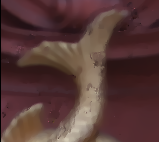} &
\includegraphics[width=0.25\columnwidth]{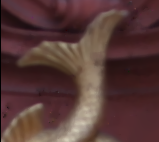} \\
&&&\\
\includegraphics[width=0.25\columnwidth]{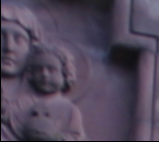} &
\includegraphics[width=0.25\columnwidth]{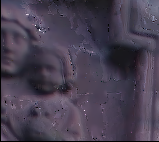} &
\includegraphics[width=0.25\columnwidth]{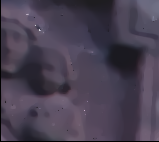} &
\includegraphics[width=0.25\columnwidth]{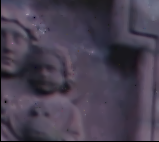} \\
&&&\\
\includegraphics[width=0.25\columnwidth]{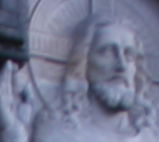} &
\includegraphics[width=0.25\columnwidth]{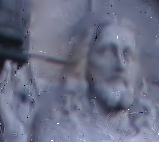} &
\includegraphics[width=0.25\columnwidth]{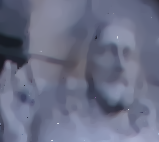} &
\includegraphics[width=0.25\columnwidth]{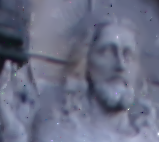} \\
&&&\\
image de référence &$\alpha = 1.0, \gamma = 0.0,$ & $\alpha = 1.0, \gamma = 0.0, $& $\alpha = 0.1, \gamma = 1.0, $ \\
 &$\lambda = 0.002 $&$ \lambda = 0.003 $ &$ \lambda = 0.002 $\\
&&&\\
\end{tabular}
\end{figure}

\begin{table*}
  \centering\footnotesize
  \begin{tabular}{ | c || c | c | c | }
    \hline
    & $\alpha = 1.0, \gamma = 0.0,$ & $\alpha = 1.0, \gamma = 0.0,$ & $\alpha = 0.1, \gamma = 1.0,$ \\
    & $\lambda = 0.002$ & $\lambda = 0.003$ & $\lambda = 0.002$ \\
    & ~\cite{pujades_bayesian_2014} & ~\cite{pujades_bayesian_2014} & (notre méthode) \\
    & ~\cite{wanner_spatial_2012} & ~\cite{wanner_spatial_2012} & \\
    \hline \hline
    \emph{fountain -- view 2} & 21.03 \quad 132 & 21.09 \quad 120 & \textbf{21.16} \quad \textbf{107} \\
    \hline
    \emph{fountain -- view 5} & 26.00 \quad 74 & 26.14 \quad 64 & \textbf{26.36} \quad \textbf{51} \\
    \hline
    \emph{fountain -- view 8} & 22.00 \quad 140 & 22.08 \quad 125 & \textbf{22.16} \quad \textbf{111} \\
    \hline
    \hline
    \emph{herzjesu -- view 2} & 21.73 \quad 186 & \textbf{21.96} \quad 153 & 21.93 \quad \textbf{143} \\
    \hline
    \emph{herzjesu -- view 4} & 23.13 \quad 194 & 23.81 \quad 130 & \textbf{23.90} \quad \textbf{115} \\
    \hline
    \emph{herzjesu -- view 6} & 18.08 \quad 349 & 18.26 \quad 287 & \textbf{18.31} \quad \textbf{273} \\
    \hline
        \hline
    \emph{charce -- view 4} & 13.74 \quad 905 & \textbf{13.85} \quad \textbf{885} & 13.73 \quad 901 \\
    \hline
    \emph{charce -- view 6} & 9.953 \quad 1360 & \textbf{10.04} \quad \textbf{1317} & 10.00 \quad 1352 \\
    \hline
    \emph{charce -- view 11} & 14.20 \quad 859 & \textbf{14.38} \quad \textbf{793} & 14.20 \quad 843 \\
    \hline
        \hline
    \emph{lion -- view 1} & 24.40 \quad 190 & 24.36 \quad 198 & \textbf{24.46} \quad \textbf{175} \\
    \hline
    \emph{lion -- view 3} & 29.16 \quad 103 & 29.15 \quad 107 & \textbf{29.57} \quad \textbf{87} \\
    \hline
    \emph{lion -- view 5} & 22.18 \quad 298 & 22.25 \quad 294 & \textbf{22.27} \quad \textbf{277} \\
    \hline
  \end{tabular}
  \caption{Résultats numériques sur des bases de données non structurées. Notre méthode est comparée aux méthodes de l'état de l'art \protect\citeNP{pujades_bayesian_2014, wanner_spatial_2012}, pour lesquelles il n'y a pas de contraintes sur les gradients de l'image ($\gamma = 0.0$). Pour chaque résultat, la première valeur est le PSNR (plus il est élevé mieux c'est), la seconde étant le DSSIM (plus il est faible mieux c'est). $\mathrm{DSSIM} = 10^4(1 - \mathrm{SSIM})$. La meilleure valeur est notée en gras.} 
  
  \label{tab:results_unstructured}
\end{table*}

Les expériences suivantes (figure~\ref{fig:exper}) ont été réalisées sur des vues réelles prises de la base de données de \citeNP{strecha_benchmarking_2008}, \emph{fountain} et \emph{herzjesu}, ainsi que sur nos propres jeux de données \emph{charce} et \emph{lion}. Les résultats numériques du tableau \ref{tab:results_unstructured} viennent confirmer les figures d'illustration en terme d'efficacité de l'ajout du terme portant sur les gradients des images.

Comme le système est faiblement contraint, des hautes fréquences apparaissent dans les zones visibles depuis peu de caméras. Ces artefacts sont accentués par une estimation très bruitée de la profondeur près des régions d'occultation (autour du poisson de la fontaine, ou du bas-relief de Jésus). Le paramètre $\lambda$ contrôlant le terme de régularisation \emph{Total Variation} a été augmenté à 0.003 pour réduire l'apparition de ces hautes fréquences. Mais le résultat est peu convainquant car les images perdent alors du détail comparées aux originales. Nous avons alors baissé $\lambda$ pour conserver tous les traits du bas-relief et ajouté le terme sur les gradients ($\gamma = 1.0$). Nous pouvons voir sur les images, quelle que soit la géométrie utilisée pour le rendu, que les artefacts disparaissent tout en préservant les détails de l'image. Le terme sur les couleurs est conservé pour que la couleur originale des images ne soit pas affectée mais mis à une valeur faible ($\alpha = 0.1$). L'ajout du terme sur les gradients a en outre permis d'empêcher l'apparition de faux contours près des frontières de visibilité, garantissant ainsi la propriété de \emph{continuité}.

Les résultats numériques mettent en évidence que l'amélioration de la qualité des images synthétisées sauf dans le cas du jeu de données \emph{charce}. En effet les meilleurs résultats sont obtenus avec un fort terme de lissage, alors que l'ajout du terme sur les gradients n'influe quasiment pas. Précisons que la reconstruction 3D de cette scène est très mauvaise du fait de la diversité des points de vues utilisés et de la présence du ciel, arrière-plan non texturé. Par conséquent de nombreuses zones ne sont pas reconstruites, et l'ensemble des pixels sur lesquels nous pouvons évaluer la qualité des résultats (l'ensemble des pixels reconstruits) est trop restreint pour que l'évaluation soit pertinente. En outre on peut penser que dans le cas où la reconstruction est extrêmement bruitée, une forte stabilisation par le terme de lissage est plus bénéfique que l'ajout du terme d'attache aux gradients. 

\section{Discussion et conclusion}
\label{sec:conclusion}

Nous avons présenté une méthode de rendu basé image qui permet de générer une nouvelle vue à partir d'un ensemble générique et non structuré d'images. Cette méthode est inspirée par les travaux de \citeNP{pujades_bayesian_2014}, qui ont oeuvré pour formaliser la plupart des << propriétés désirables >> listées dans l'article phare de \citeNP{buehler_unstructured_2001}. Leur approche fut d'introduire une formulation bayésienne du problème de rendu et d'obtenir la vue cherchée par un processus d'optimisation. La seule propriété qu'ils n'ont pu formaliser fut la propriété de \emph{continuité}, qui énonce que les contributions de chaque vue source doivent être des fonctions continues des coordonnées des pixels.

Nous avons montré qu'un moyen de garantir cette \emph{continuité} est de déclarer que les contours, textures et détails ne devraient pas être créés dans l'image cible s'ils ne sont pas présents dans les images sources aux endroits visibles. Cela implique l'ajout d'un terme additionnel portant sur les données sources, basé sur les gradients des images. L'énergie ainsi modifiée peut être minimisée en résolvant de façon itérative un système linéaire dérivé de la fonctionnelle. Ce système est alors plus contraint et mieux conditionné que le précédent, ce qui empêche l'apparition d'artefacts près des frontières de visibilité. Ce résultat montre une nette amélioration par rapport aux précédentes méthodes de rendu basées sur les intensités, à la fois en terme de mesure qualitative et en terme de qualité subjective. 

Cette méthode pourrait être retravaillée pour optimiser directement les gradients de la vue cible, plutôt que les intensités~; puis l'intensité de l'image pourrait être reconstruite en résolvant l'équation de Poisson, comme il est fait par \citeNP{kopf_image-based_2013}. Cela devrait complètement enlever toutes les variations dans l'image synthétisée qui viennent de discontinuités des fonctions de visibilité, qui sont toujours visibles dans nos résultats, bien qu'atténuées.

\acknowledgements{Nous remercions la DGA pour nous avoir co-financé sur ce travail de recherche.}

\bibliography{TS2016}

\end{document}